\documentclass[letterpaper,twocolumn,10pt]{article}
\usepackage{usenix-2020-09}

\usepackage{times}
\usepackage{mathtools}
\usepackage{titling}
\usepackage{enumitem}
\usepackage[normalem]{ulem} 
\usepackage[export]{adjustbox}

\usepackage{titlesec}
\titlespacing*\section{0pt}{6pt plus 0pt minus 0pt}{2pt plus 0pt minus 0pt}
\titlespacing*\subsection{0pt}{6pt plus 0pt minus 0pt}{2pt plus 0pt minus 0pt}
\titlespacing*\subsubsection{0pt}{6pt plus 0pt minus 0pt}{2pt plus 0pt minus 0pt}

\usepackage{algorithm}
\usepackage[noend]{algpseudocode} 
\usepackage{algorithmicx}

\usepackage{etoolbox}
\usepackage{tikz}
\usetikzlibrary{tikzmark}
\usetikzlibrary{calc}

\newcommand{\ALGtikzmarkcolor}{black}
\newcommand{\ALGtikzmarkextraindent}{4pt}
\newcommand{\ALGtikzmarkverticaloffsetstart}{-.5ex}
\newcommand{\ALGtikzmarkverticaloffsetend}{-.5ex}
\makeatletter
\newcounter{ALG@tikzmark@tempcnta}

\newcommand\ALG@tikzmark@start{%
	\global\let\ALG@tikzmark@last\ALG@tikzmark@starttext%
	\expandafter\edef\csname ALG@tikzmark@\theALG@nested\endcsname{\theALG@tikzmark@tempcnta}%
	\tikzmark{ALG@tikzmark@start@\csname ALG@tikzmark@\theALG@nested\endcsname}%
	\addtocounter{ALG@tikzmark@tempcnta}{1}%
}

\def\ALG@tikzmark@starttext{start}
\newcommand\ALG@tikzmark@end{%
	\ifx\ALG@tikzmark@last\ALG@tikzmark@starttext
	\else
	\tikzmark{ALG@tikzmark@end@\csname ALG@tikzmark@\theALG@nested\endcsname}%
	\tikz[overlay,remember picture] \draw[\ALGtikzmarkcolor] let \p{S}=($(pic cs:ALG@tikzmark@start@\csname ALG@tikzmark@\theALG@nested\endcsname)+(\ALGtikzmarkextraindent,\ALGtikzmarkverticaloffsetstart)$), \p{E}=($(pic cs:ALG@tikzmark@end@\csname ALG@tikzmark@\theALG@nested\endcsname)+(\ALGtikzmarkextraindent,\ALGtikzmarkverticaloffsetend)$) in (\x{S},\y{S})--(\x{S},\y{E});%
	\fi
	\gdef\ALG@tikzmark@last{end}%
}

\apptocmd{\ALG@beginblock}{\ALG@tikzmark@start}{}{\errmessage{failed to patch}}
\pretocmd{\ALG@endblock}{\ALG@tikzmark@end}{}{\errmessage{failed to patch}}

\usepackage{caption}

\usepackage{cleveref}

\crefformat{section}{\S#2#1#3} 
\crefformat{subsection}{\S#2#1#3}
\crefformat{subsubsection}{\S#2#1#3}

\newcommand{\graphpy}{\textsc{GraphPy}}

\usepackage{listings}
\usepackage{xcolor}

\definecolor{codegreen}{rgb}{0,0.6,0}
\definecolor{codegray}{rgb}{0.5,0.5,0.5}
\definecolor{codepurple}{rgb}{0.58,0,0.82}
\definecolor{backcolour}{rgb}{0.95,0.95,0.92}

\lstdefinestyle{mystyle}{
  commentstyle=\color{codegreen},
  keywordstyle=\color{magenta},
  numberstyle=\tiny\color{codegray},
  stringstyle=\color{codepurple},
  basicstyle=\ttfamily\footnotesize,
  breakatwhitespace=false,         
  breaklines=true,                 
  captionpos=b,                    
  keepspaces=true,                 
  numbers=left,                    
  numbersep=5pt,                  
  showspaces=false,                
  showstringspaces=false,
  showtabs=false,                  
  tabsize=2
}
\usepackage{authblk}
\usepackage{balance}

\begin{document}
%

\title{\Large \bf Single-GPU GNN Systems: Traps and Pitfalls}


\author{Yidong Gong}
\author{Arnab Tarafder}
\author{Saima Afrin}
\author{Pradeep Kumar}
\affil{William \& Mary}


%


\date{}
\maketitle
\thispagestyle{empty}

%
\begin{abstract}
The current graph neural network (GNN) systems have established a clear trend of not showing training accuracy results, and directly or indirectly relying on smaller datasets for evaluations majorly. Our in-depth analysis shows that it leads to a chain of pitfalls in the system design and evaluation process, questioning the practicality of many of the proposed system optimizations, and affecting conclusions and lessons learned. 
We analyze many single-GPU systems and show the fundamental impact of these pitfalls. We further develop hypotheses, recommendations, and evaluation methodologies, and provide future directions. Finally, a new reference system is developed to establish a new line of optimizations rooted in solving the system-design pitfalls efficiently and practically. The proposed design can productively be integrated into prior works, thereby truly advancing the state-of-the-art.

\end{abstract}

\section{Introduction} \label{sec.intro}
Many applications have become data-driven today, where deep learning (DL) has gained prominence. 
Within this ecosystem, many real-world data can be stored as \textit{sparse matrices} or \textit{graphs}.
To this end, graph neural network (GNN) models, e.g., GCN~\cite{gcn17iclr}, GAT~\cite{gat18iclr}, GIN~\cite{xu2019powerful}, and others~\cite{gatedgraph2017,zhang18, graphsage17} are playing an increasingly important role. 
System optimizations play a critical role in improving the training time, which has become exceedingly time-consuming even when using powerful accelerators, such as GPUs.
To this end, several GNN systems~\cite{wang2021gnnadvisor,huang2021understanding,gespmmsc2020,fu2022tlpgnn,wu2021seastar,chen2020fusegnn,zhang2022understanding,wang2023tc,ye2023sparsetir} have proposed numerous system-level optimizations and reportedly led up to 15$\times$ speedup in training runtime in single-GPU.

\begin{figure*}[t]
 \centering
  \includegraphics[scale= 0.47]{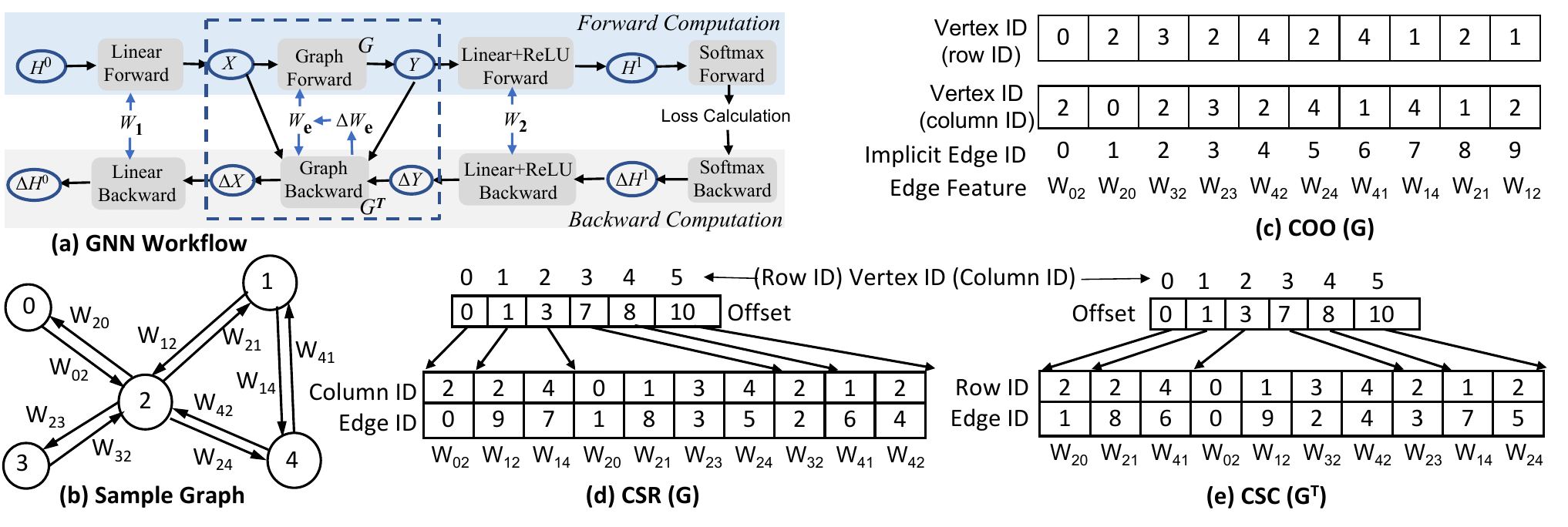}
  \vspace{-12pt}
  \caption{\small (a) A GNN layer: X and Y are activation tensors, and $W_e$ is the edge-level tensor of the sparse kernel, e.g., attention score in GAT. They are also \textit{state tensor}, as they are used in the backward computation. $G$ is a sparse matrix, and $G^T$ its transpose. (b) Sample graph/sparse matrix (c, d, and e) CSR, CSC, and COO representation along with edge ID indirection. \textit{Note:} {DGL} uses this storage design where consecutive edge ID is used in COO (as shown). Therefore, CSR and CSC store explicit edge ID arrays. The storage format is discussed in-depth in \cref{sec.mem}.}
  \label{fig-format}
  \vspace{-10pt}
\end{figure*}

In this work, we analyze such system-level optimizations and highlight that they indicate an inadequate understanding of the unique need for DL/GNN computation. \textit{Forward computation} calls many \textit{kernels} (operations) from the input layer to the output layer deriving prediction values. They are compared with the ground truth to calculate the loss.
\textit{Backward computation} uses the loss and invokes kernels from the output layer to the input layer to derive gradients and update the model parameters.
The backward computation accesses \textit{state-tensor}-- results that are produced in the forward computation (maybe \textit{sparse}) and may require a \textit{transpose} (Fig.~\ref{fig-format}a).

\begin{figure}[t]
  \centering
  \includegraphics[scale= 0.4,center]{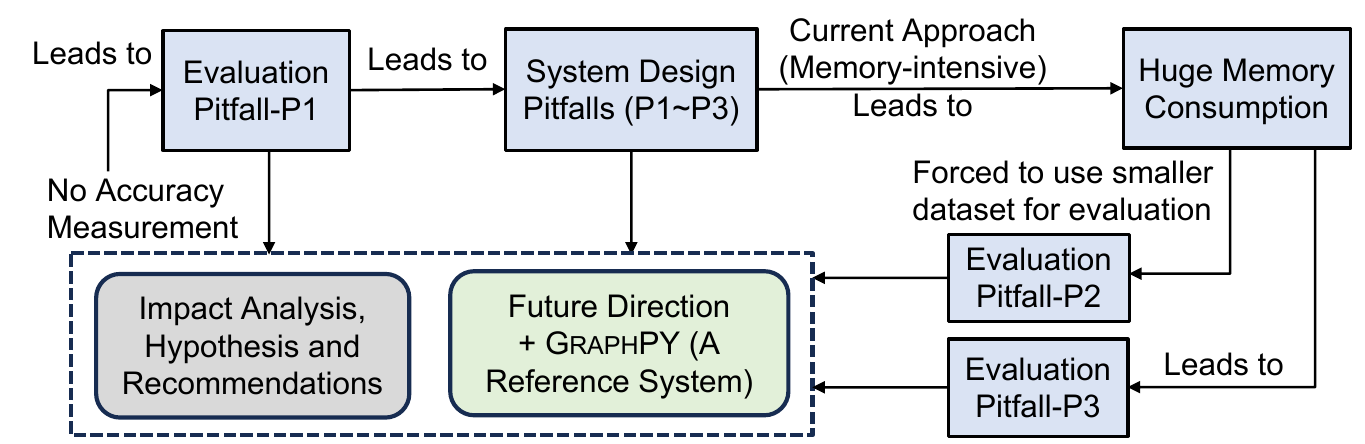}
  \vspace{-12pt}
  \caption{\small The complex relationship between pitfalls, and our recommendation. {\graphpy} is the proposed reference system.}
  \label{fig-pitfall}
\end{figure}

These unique requirements differentiate the GNN computation paradigm from forward-only related fields, e.g., graph analytics.
Hence when it is not understood correctly, existing GNN systems suffer from many pitfalls in \textit{evaluation and system design}. 
To illustrate, the majority of single-GPU GNN systems have established a clear trend of \textit{no training accuracy measurement}~\cite{huang2021understanding, wang2021gnnadvisor, zhang2022understanding, wu2021seastar, wang2023tc, gespmmsc2020,fu2022tlpgnn,wang2021flexgraph,hu2021efficient,jangda2021accelerating, liang2020engn, zhang2021boostgcn, kim2022analyzing, yan2020hygcn,ye2023sparsetir}, while a few do not even implement backward computation. This has undermined the evaluation process and has set off a chain of pitfalls as shown in Fig.~\ref{fig-pitfall}.

In the GNN system, forward and backward computations have various design trade-offs, that we study in this paper. System building and performance optimization require navigating those trade-offs. 
On the other hand, A scientific evaluation methodology is a fundamental process to support intuitions, system optimizations, and conclusions. 
Hence evaluation pitfalls break this most important feedback loop leading to a compromised system design, misunderstanding of trade-offs, over-optimistic performance gain, and affecting other aspects of GNN workflow negatively: wrong conclusions, improper lessons learned, and hindering the adoption of these research advances in academia and industry. Such pitfalls also \textit{stifle future innovations} because a system without any pitfall is unlikely to outperform such works.


This paper, for the first time, presents these pitfalls with in-depth systematic analysis, still unknown to the community. The analysis is backed by the interpretation of existing evaluation results, exploring additional evaluation methodology, conducting new experiments, code-study, and a comprehensive prevalence study. \textit{For clarification, we focus on end-to-end system building and optimizations of which individual kernel-level optimizations play a significant role in a single-GPU system.}
Contributions are listed next.

\noindent $\bullet$
\textbf{Unveiling Pitfalls.}  
\textit{No training accuracy measurement} indicates a serious evaluation pitfall, as our measurements show \textit{abnormal accuracy} in many GNN systems published in top-tier system conferences. 
Our investigation confirms the presence of \textit{system design pitfalls} and numerous \textit{implementation issues} in those systems.
DGL, a widely used baseline, suffers no such pitfalls as it deploys a memory-intensive design resulting in frequent out-of-memory during peer comparison by prior works. Hence, prior works have concluded better training time largely or exclusively using smaller datasets. However, our \textit{novel methodology on framework overhead measurement} reveals that training speedup on smaller datasets is purely due to low framework overhead and not due to individual optimizations. This observation leads to a \textit{pitfall in runtime evaluation comparison}.

\noindent  $\bullet$
\textbf{Impact Analysis, Hypothesis, and Recommendation.} 
We explain that the pitfalls are not measurement oversight or implementation issues as they have led to serious flaws in system design and evaluation; solving which requires changes in the evaluation methodology and critical design thinking respectively.
Moreover, rather than focusing excessively on the negative aspects of the
prior practices we observed in the literature, we
aim to increase awareness of the existence of these flaws through answering these questions: \textit{what} are the major implications of these pitfalls in current GNN systems (\textit{Impact}); \textit{why} do such pitfalls occur despite the best intentions of the researchers and community (\textit{Hypothesis}); and \textit{how} to enforce that future works do not suffer from such pitfalls (\textit{Recommendation}). 




\noindent $\bullet$
\textbf{Future Direction and a Reference System.}
Beyond pitfalls identification and recommendations, we outline three directions to further deal with pitfalls: a) advent of framework overhead establishes a new area of research and shows an analogy to Operating Systems research for future direction; b) a few pitfalls can only be corrected in a pervasive manner; and c) need for a reference single-GPU GNN system to serve as a baseline to advance the future research. The aim is to identify \textit{practical} and \textit{efficient} design choices that can be integrated productively into prior works to solve some pitfalls. 
In response, we introduce {\graphpy} that explores a new territory of system optimizations rooted in solving the system design pitfalls. To this end, we propose a novel \textit{edge ID reordering} and correct \textit{exploitation of dataset symmetry} to lower memory consumption and \textit{better data locality} for significant speedup.

Evaluations are done throughout the paper to show that pitfalls exist. Evaluations in \cref{sec.exp} establish {\graphpy} as a perfect reference system as it reduces memory consumption on average by 6.92$\times$ on GCN, 3.4$\times$ on GIN, and 1.96$\times$ on GAT-1 models in comparison to DGL. 
Further, {\graphpy} achieves on average 1.69$\times$, 1.22$\times$, 2.20$\times$, 1.69$\times$ speedup for GCN, GIN, GAT-1, and GAT-3 compared to DGL on mid-size datasets. {\graphpy} can perform much better (1.16$\times$--2.87$\times$) than optimized and often workload-balanced kernels of FeatGraph~\cite{featgraphsc20}, GE-SpMM~\cite{gespmmsc2020}, GNNAdvisor~\cite{wang2021gnnadvisor}, TC-GNN~\cite{wang2023tc}, Huang et al~\cite{huang2021understanding}, CuSparse, and DGL.
More importantly, the proposed baseline can train GCN on a billion-edge graph on a single GPU due to memory saving while prior works cannot train even a 500 million-edge graph as DGL, Pytorch-Geometric~\cite{pytorchg2019} (PyG) show out-of-memory state.

Our recommended evaluation strategy shows that pitfalls-infested systems overstate the performance gain, and a few are even slower than the same baseline on mid-size datasets.  
Furthermore, pitfalls impact future evaluations as well, by showing performance evaluation of {\graphpy} with a pitfall-infested system, showing that pitfalls stifle future innovations. 




\noindent \textbf{Importance.}
Related to our observations are works that uncover pitfalls in \textit{application} aspects of DL or ML (machine learning). 
A distinguished paper award recipient~\cite{doanddont2022and} in USENIX security, 2022, systematically describes the ML pitfalls in the security domain.
{Earlier, another best paper awards recipient~\cite{arewereally2019we} explains similar pitfalls in the recommendation system}.
These papers show that pitfalls are more commonplace in the DL/ML field and their studies are valued by computer science communities. 
Different from them, our work shows the system-specific pitfalls, where single-GPU GNN systems use the same (or minor changes to) model specifications and datasets but a new back-end. 

Moreover, GNN optimizations have wider applications in exploiting sparsity in the broader DL community, where models are either pruned~\cite{han2015learning,han2016iclr,sparsetransformer2019,frankle2018lottery} to make its layer sparse that requires re-training or can be defined initially~\cite{liu2018rethinking} as sparse thereby requiring similar to GNN sparse kernels. Hence pitfalls in GNN systems will likely impact this area as well.

The paper is organized as follows. We present prevalence, scope and clarification in \cref{sec.prevalence},  background in \cref{sec.background}, pitfalls and related works in \cref{sec.accuracy} and \cref{sec.evaluation}. Future direction and {\graphpy} is discussed in \cref{sec.analyze}, evaluations in \cref{sec.exp}, and conclusion in \cref{sec.conclusion}.

\section{Prevalence Analysis and Scope} \label{sec.prevalence}

In this work, we focus on \textit{single-GPU GNN systems}, and analyze in-depth over 20 systems~\cite{wang2021gnnadvisor, fu2022tlpgnn, huang2021understanding, gespmmsc2020, zhang2022understanding, featgraphsc20, dgl2019, jangda2021accelerating, liang2020engn, yan2020hygcn, pytorchg2019, chen2020fusegnn,tian2020pcgcn,wu2021seastar,wang2023tc,kim2022analyzing, yan2020characterizing, tian2020pcgcn,baruah2021gnnmark, ye2023sparsetir, waleffe2023mariusgnn} in the last 4 years, and are from the top systems and HPC conferences, such as OSDI, ASPLOS, EuroSys, ATC, MLSys, SC, IPDPS, HPDC, and PPoPP among others; hence should not be termed as cherry-picking. 
We analyze and comment on the training process as a whole, which automatically focuses on system building, of which individual kernel optimizations play a big part in a single-GPU system. It is backed by a detailed code study and evaluation of many open-sourced works.
In addition, we would like to make the following important points:

\begin{itemize}[noitemsep,topsep=0pt]
 
\item The paper is not related to the reproduction issue. \textit{We indeed can observe and reproduce similar results}. The points raised are based on our interpretation of those results, and additional evaluations that we do on accuracy, overhead, and kernel run-time measurements. 


\item This paper aims to identify common pitfalls inherent in many GNN systems and elucidate their impact on research. It is possible that a correct single-GPU GNN system, in case our extensive search missed them, does not have these pitfalls. But that does not solve the case that these pitfalls are prevalent within the current landscape.
It is essential to note that not all papers within the GNN domain succumb to every identified pitfall. This paper does not seek to cast blame on all techniques proposed by existing works. Instead, our objective is to uncover general pitfalls that, if unaddressed, can hinder the advancement of GNN system research.

\item  
Distributed GNN systems or CPU-GPU systems~\cite{wang2021flexgraph,cai2021dgcl, md2021distgnn, zheng2020distdgl, p3gandhi2021,tripathy2020reducing, wang2021flexgraph, jia2020improving, Thorpe2021DorylusAS, ByteGNN22, DSP23, Yang2022WholeGraphAF, yang2022gnnlab,mgg2023} are not in the scope. 
This way, the role of kernel optimizations and their cost of integration with a single-GPU system can be analyzed in-depth without worrying about the costs of CPU-GPU and inter-GPU communication. 
Also, with increasing memory size in modern GPUs,  and as the performance and memory consumption of a single-GPU system forms the basis for a distributed design, analyzing the former is very important. 

\item
Inference-oriented works~\cite{geng2020awb, zhang2021boostgcn, yan2020characterizing} are not in scope.

\end{itemize}









\section{Background on GNN Computation} \label{sec.background}


\noindent \textbf{Graph/Sparse Matrix Storage Formats.}
A graph is represented as G = (V, E), where V and E refer to the vertex/row set and the edges/non-zero elements respectively. 
We continue to use both the graph and sparse linear algebra terminologies. Specifically, features and computation are referred to as vertex-level and edge-level, while rows, columns, and non-zero elements are used to refer to datasets.

Fig.~\ref{fig-format} shows an example of a sample graph and its corresponding representations. 
The \textit{compressed sparse row}(CSR) format stores non-zero elements in a row sequentially and uses the offset array to point to the start of the row. The reverse graph or transpose in the case of a directed graph is also needed to be stored; in this case, CSR stores the rows consecutively, while \textit{compressed sparse column} (CSC) format stores the columns consecutively. The \textit{degree} of a row is the row length. The figure also shows an edge ID array as part of the storage format which is discussed later.

\noindent \textbf{Taxonomy of GNN models: }
Here we establish a taxonomy of classifying GNN models into broad categories. Each category is associated with two types of graph-related kernels~\cite{dgl2019} called \textit{gSpMM} (generalized sparse matrix dense matrix multiplication), and \textit{gSDDMM} (generalized sampled dense dense matrix multiplication).
gSpMM has many variations, including the weighted and unweighted versions, uses various reduction operations (e.g. sum, min, max, etc.).
The gSDDMM kernel can be thought of as a derivation of edge-level tensors  either from vertex-level tensors or from a vertex-level tensor and an edge-level tensor. 
It also has variations due to various operations (sum, division, etc.) that need to be performed.

$\bullet$
\textbf{Class A: Vertex-Level + Edge- Level Tensors.}
The GNN models, such as GAT, RGAT, GaAN~\cite{gat18iclr, rgcn2018, zhang18}, etc. use the weighted sum aggregation in its gSpMM, called \textbf{gSpMMve}, where the last two letters signify the vertex-level and edge-level input tensor types. The edge-level tensors are trainable. 
We also need another gSpMM variant, called \textbf{gSpMMe}, and two variants of gSDDMM. In \textbf{gSDDMMvv}, the vertex-level features of row ID and column ID of each non-zero element of the sparse matrix perform dot product to generate edge-level tensors, while in \textbf{gSDDMMve}, vertex-level and edge-level tensors are accessed using the graph to arrive at the resultant edge-level tensor.

$\bullet$
\textbf{Class B: Only Vertex-Level Tensors.}
The GNN models such as GCN\cite{gcn17iclr} and GIN\cite{xu2019powerful}, use an unweighted sum to aggregate vertex-level neighbor features, Hence, it only needs vertex-level tensor, and not edge-level tensor. This kernel is called \textbf{gSpMMv}, a variant of gSpMM.
Another related kernel is \textit{normalization by degree} of vertex-level tensors, which is needed in GCN.


\noindent \textbf{Backward Conputation.}
For gSpMMve ($Y \leftarrow AX$), its backward computation needs to include both the \textit{transposed gSpMMve} ($\delta X \leftarrow A^T \delta Y$) and gSDDMMvv ($\delta W_e \leftarrow A \odot (X^T \triangle Y)$), where $\delta$ represents gradients of the tensors. We denote transposed gSpMMve as  $gSpMMve^T$.

\section{Missing Accuracy \& System Design Pitfalls} \label{sec.accuracy}




\noindent
\textbf{Introduction and Prevalence Study.}
Several GNN systems~\cite{huang2021understanding, wang2021gnnadvisor, zhang2022understanding, wu2021seastar, wang2023tc, gespmmsc2020,fu2022tlpgnn,wang2021flexgraph,hu2021efficient,jangda2021accelerating, liang2020engn, zhang2021boostgcn, kim2022analyzing, yan2020hygcn,ye2023sparsetir} have established a clear trend of not reporting training accuracy in their paper. 
A few of them do not even have a backward computation and rely on an emulated version of forward computation, which cannot fit into any training process. This pitfall questions the validity of the evaluation process as we show that many influential papers have abnormal accuracy, which is due to requirement misunderstanding that leads to fundamental issues in their system design.

\subsection{EVAL-P1: No Accuracy Measurement}

We measure training accuracy if prior works have backward computation implementation and have not reported accuracy.  Datasets are discussed in \cref{sec.exp}, and experiments are run on an Nvidia A100 GPU.
We only picked top-tier conference papers to do the measurements.
Also, see \cref{sec.accuracy.discuss} for more discussion.



\begin{figure}[t]
  \includegraphics[scale= 0.47]{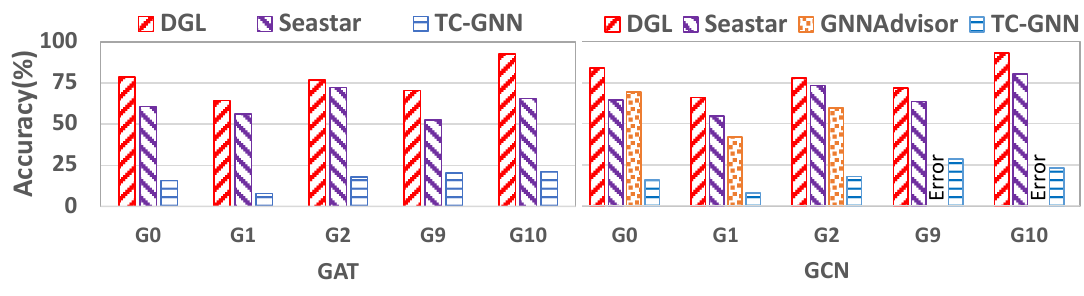}
  \vspace{-18pt}
  \caption{\small Accuracy comparison: TC-GNN does not provide GAT, hence AGNN is substituted here as both are attention-based(Class A GNN). We also discuss other works in the text. DGL is for reference. } 
  \label{pitfall-accuracy}
\end{figure}

\noindent \textbf{Accuracy measurement result} is shown in Fig.~\ref{pitfall-accuracy}. DGL accuracy is for reference. Observation is discussed next.



\textbf{Class A GNN.}
Seastar~\cite{wu2021seastar} shows a drop in accuracy by 4.5\% - 26.9\%, while TC-GNN~\cite{wang2023tc} have very very low accuracy on all the datasets for GAT. 
FuseGNN~\cite{chen2020fusegnn} GAT shows NaN during training on Reddit and could achieve only 58\% accuracy, way below DGL's 92.4\%.
We did not run Huang et al~\cite{huang2021understanding}, and TLPGNN~\cite{fu2022tlpgnn} that have GAT but they do not implement backward computation. 
GNNAdvisor does not provide GAT implementation.


\textbf{Class B GNN.}
GCN training accuracy in TC-GNN is abnormally low. Seastar accuracy drops by 4.5\% - 19.8\%. 
GNNAdvisor~\cite{wang2021gnnadvisor} drops by 14.8\% -- 24\%. Also, GNNAdvisor throws a memory corruption error for Reddit and OGB-product, which is also reported by others~\cite{fu2022tlpgnn}.
Huang et al and TLPGNN do not have backward computation implemented.  
Ge-SpMM~\cite{gespmmsc2020} which does not report accuracy, could not be run due to major code reorganization in DGL to which it was integrated. 
Featgraph~\cite{featgraphsc20}, which is integrated into DGL, we could not install it correctly despite help from its authors and trying several GitHub branches though it did report accuracy on Reddit. Except for DGL, only PyG and FuseGNN have normal accuracy on GCN.


\noindent
\textbf{Fundamental Problem or Evaluation Oversight?}
Academic prototypes usually do not have the same level of evaluation as commercial prototypes. So one can claim the missing accuracy is simply an evaluation oversight. 
However, any significant drop in accuracy cannot be ignored as these systems usually aim to keep the same GNN model as the baseline (DGL in this case), with minor changes specifically for advanced optimizations such as operator (or kernel) fusion. 
Hence significant errors in accuracy in those same datasets for which systems are evaluated cannot be termed as an oversight. As discussed in \cref{sec.semantic}, we have uncovered fundamental issues by debugging accuracy errors that can not be ignored as they also contribute to better performance. In the end, we discuss the impact, hypothesis, and recommendation together.

\subsection{System Design Pitfalls} \label{sec.semantic}
The drop in accuracy is attributed to a semantic error in understanding the requirements, which has led to the wrong system design. We analyze their open-sourced code and identify at least one reason for abnormal accuracy. 
This category contains three generic system design pitfalls as presented next.



\subsubsection{SYS-P1: Omitted State Tensor}
Many works fused the kernel to optimize the system performance. However, such works do not consider the state tensor requirement for backward computation. 

\noindent\textbf{Definition and Example.}
The backward computation of any operator needs to access the \textit{state tensor}-- the input and/or output tensors of the corresponding forward operator. Whether it needs just the input tensor or output tensor is operator-specific. Such tensors must be saved during forward computation to be used during backward computation.  
Hence, when backward computation is not implemented, their emulated forward computation does not generate and materialize many state tensors. Hence, kernels could be fused across operators by writing some computations to shared memory or even to device registers rather than to global memory to gain performance advantage.
Without considering the data write and data read to global memory for the state tensor(s), their performance speed-up is over-stated.



Kindly note that state tensors requirement should not be seen from sparse operations only, but many other operations, such as ReLU, Dropout, etc. operations need to store state tensors.  
Huang et al~\cite{huang2021understanding} and TLPGNN~\cite{fu2022tlpgnn} prototyped one giant fused forward kernel that fused many sparse kernels along with other operations, such as ReLU, etc. 
Hence, their forward-only implementation was able to take advantage of kernel fusion to completely remove the computation and storage of many state tensors. {TLPGNN did not materialize the attention score in GAT-- an edge-level state tensor, while Huang et al. and TLPGNN did not materialize the state-tensor for the ReLU operation.}
Kindly note these two works claim to optimize GNN training, and not GNN inference, hence, they are not analyzed for the latter. 


\noindent $\bullet$
\textbf{Fundamental Issues or Implementation Issues?}
By substituting emulated forward computation as a replacement for GNN training, such systems over-state the performance gain and this is not a good benchmarking practice. 
In other words, the speedup claimed by such works is not practical even if their kernel optimizations are valid.
Indeed, \textit{kernel fusion}
is a well-studied technique to improve performance in many scientific and graph computations in GPUs. However, in deep learning, kernel fusion is not so straightforward where 
materialization of the state tensor is a fundamental design factor that needs to be considered carefully.
The existing optimizations will not hold if someone decides to fix this pitfall. Further, fixes need critical thinking and design changes, such as the re-computation of state-tensors~\cite{zhang2022understanding} which itself requires saving some other low-cost tensors as state-tensors. This saving and re-computation procedure leads to extra slowdown.

\subsubsection{SYS-P2: Missing Sparse Matrix Transpose}
\label{sec-P2}

A few GNN systems did not transpose the edge-level state tensor during backward computation leading to unrealistic and non-practical performance gain.

\noindent\textbf{Definition and Example.}
Backward gSpMM computation ($\delta X \leftarrow A^T \delta Y$), requires transpose of the edge-level state tensors($A$), the sparse matrix. 
The sparse matrix (A) itself can be viewed as two parts: the static part(the graph topology) and the dynamic part(the edge-level state tensor which represents the edge values). While, one can pre-process the graph to keep the transpose of the topology, the edge-level tensor requires transpose at run-time.
While many works~\cite{fu2022tlpgnn,huang2021understanding} do not implement backward computation by ignoring this problem, TC-GNN~\cite{wang2023tc} substitute forward gSpMM in place of backward gSpMM for its attention-based GNN. 


Kindly note that the sparse matrix (the dataset) is square in the case of GNN, and hence not performing transpose does not lead to any dimension mismatch, and cannot be caught during compilation or runtime unless accuracy is measured. 


\noindent
\textbf{Fundamental Issues or Implementation Issues?}
Backward gSpMM is fundamentally different than forward gSpMM because the transpose need introduces a basic trade-off. 
To illustrate, prior works assume the sparse matrix (static + dynamic parts) in a row-major layout, where forward gSpMM reads the row using coalesced memory access, though their finer details vary.
In transposed gSpMM, if one does not perform an explicit transpose or alter the algorithm, the sparse matrix now needs to be read in column order sacrificing the coalesced memory access. Even if one keeps the transposed topology, reading the row-major edge-level tensor in column order is needed.
\textit{It is not possible to read a row-major sparse matrix in column order without additional metadata or processing.} 




Currently known solutions are inefficient as discussed in \cref{sec.mem}.
So, dropping the requirement itself, either through missing backward computation or substituting forward gSpMM in place of backward gSpMM does not present us with those trade-offs in various designs possible.
Thus, these works have unrealistic and non-practical performance gains for training.

\subsubsection{SYS-P3:Incorrect Order of Backward Operations}

A few works that attempt kernel fusion do not order their backward operation correctly, which leads to better performance at the expense of wrong training. 

\noindent\textbf{Definition and Example.}
The backward operations, as the name suggests, invoke the kernels in the reverse order to the forward operations. For example, if operator A is called first followed by operator B, then the backward computation first needs to call the backward of operator B followed by the backward of operator A. However when the operators A and B are fused in forward, their backward requires a different custom fused kernel to take care of the ordering.

For example, GNNAdvisor~\cite{wang2021gnnadvisor} fuses the gSpMM and normalization by degree operators during the forward. However, its backward computation did not reverse the order at all. 
For Seastar~\cite{wu2021seastar} in GCN, the accuracy drop is similar to GNNAdvisor. This is because Seastar relies on a lambda function specification for fused forward SpMM and normalization by the degree in the forward() API of the GCN layer to generate CUDA code. The generated code always calls normalization after gSpMM. 
We suspect that code-generation logic may not be intelligent enough to generate correct backward fusion.  

\noindent
\textbf{Fundamental Issues or Implementation Issues?}
An independent \textit{normalization by degree} fetches the degree of every \textit{row}, thus occurring $|V|$ data load. 
It can easily be fused to forward gSpMM, as normalization happens on the output-tensor of gSpMM, which requires fetching the degree of every \textit{row}.  
However, during backward, it should be called on the input-tensor of the gSpMM. Hence, a correct fusion during backward computation needs to fetch the degree of column ID of every \textit{non-zero element}, thereby fetching $|E|$ data load.   

So when this operation is wrongly performed at the output of the backward gSpMM, the fusion wrongly fetches the degree of each row, instead of every non-zero element occurring $|V|$ data load. Hence the fused kernel gains performance due to less data load and computation. 
Clearly, fixing such issues is not simple, but requires greater scrutiny of system design.




\subsubsection{Other Errors}
The following additional errors have been observed in GCN: no model bias parameter in GNNAdvisor, Huang, et al, and TLPGNN in their GCN; 
and absence of bias and \textit{normalization by degree} array in TC-GNN in its GCN. In DGL, the bias parameter is enabled by default, while performing normalization by degree operation.

In GAT, no dropout layer is there in the fused GAT model of Seastar, TLPGNN, and Huang et al., introducing which requires complex changes to their backend due to the fusion.

\subsection{Discussion} \label{sec.accuracy.discuss}
The accuracy evaluation is a very time-consuming exercise for systems that are not maintained, hence we limited accuracy evaluation to top-tier conferences only. This is because of various compilation, installation, and usage issues that are common with academic prototypes. Additional supporting software has since changed substantially, like deprecated APIs. 
Many systems rely on generated features and labels, while each system has its own way of loading the dataset, which takes time to understand. And we modified their training script to read it from the labeled dataset. 
Further, our goal is to be fully sure when we report abnormal accuracy. 

It is possible that other systems that did not show accuracy may have correct accuracy. However, this fact may not solve the pitfall as future works may continue to make erroneous decisions if accuracy is not measured. 



\noindent
\textbf{Impact: Questionable Speedup.}
Lack of accuracy measurement breaks an important feedback loop that could have made researchers aware of possible issues in the system design. The presence of such issues questions whether the claimed speedup is practical or not. In systems, many optimizations are due to better insights into the fundamental trade-offs. Hence, when evaluations are not thorough, the performance speedup result could be misleading as we show in \cref{sec.exp.kernel}. 

\noindent
\textbf{Impact: Stifling Future Innovation.}
A new system with valid optimization will find it very hard to outperform such works without committing similar or other pitfalls. Indeed, it is very hard to find a working single-GPU GNN system as a baseline except DGL which can support \textit{both the GNN classes}. Hence if the community is not aware of them, genuine progress in system optimization may not happen.  

\noindent \textbf{Hypothesis: Automatic Backward Computation.}
We do not doubt the intention of the researcher for the pitfalls, because in that case, they would have not made their code open-source.
We hypothesize that a lack of understanding of the DL framework abstraction could be one possible reason: the model and layer definitions only explain the forward computation, while the backward code is invoked automatically for each forward operation and remains hidden from the usual code walk-through. 
Hence, many works have shown gSpMM performance comparison, but we could not find any GNN system that has specifically compared gSpMM$^T$, which appears only in the backward computation. 
However, such low-level verification needs to be extended to fused kernels, backward kernels, etc. though one cannot verify the backward operations semantically, which is related to requirement understanding such as the order of kernels in fused backward, etc. 
Hence end-to-end verification such as accuracy measurement is needed. 



\noindent \textbf{Recommendation \#1.}
It is aptly clear that measuring training accuracy is an important feedback loop about the correctness of design decisions which the majority of single-GPU GNN systems avoid.
and we humbly request the community and researcher to enforce this recommendation together so that the proposed optimization can be justified for the advancement of the systems research.

\noindent \textbf{Recommendation \#2.}
For the systems with no backward implementation or with accuracy issues, we believe that directly evaluating their kernel can be another methodology to make sure that we do not discard their advances in basic kernel design, like gSpMM and gSDDMM. This may not be possible if prior works have only fused kernels with pitfalls, and hence they risked being sidelined for evaluations by future works. 

\section{Framework Overhead \& Memory Usage} \label{sec.evaluation}





\noindent \textbf{Introduction and Prevalence Study.}
The end-to-end training time in small datasets is dominated by framework overhead instead of kernel runtime. So, the training speedup in such cases is due to the lower framework overhead (unknowingly) instead of better kernel runtime (claimed). 

Almost all single-GPU GNN training systems~\cite{wang2021gnnadvisor,fu2022tlpgnn,chen2020fusegnn,gespmmsc2020,dgl2019,wu2021seastar,tian2020pcgcn,zhang2022understanding,hu2021efficient,wang2023tc,kim2022analyzing, yan2020characterizing, tian2020pcgcn,baruah2021gnnmark,ye2023sparsetir} 
are impacted by this pitfall, as they exclusively or majorly relied on smaller datasets to conclude better training runtime. 
On the other hand, the results on mid-size datasets are scant, or inconclusive as these works have frequently shown that DGL runs out of memory on GNN training on mid-size datasets (\cref{sec.mem}), such as Reddit, a widely used labeled mid-size dataset. To be more specific,
GNNAdvisor, Ge-SpMM, TC-GNN, and others~\cite{hu2021efficient,baruah2021gnnmark} have relied exclusively on smaller datasets for training runtime comparison.
FuseGNN~\cite{chen2020fusegnn}, dgNN, and Seastar used only one mid-size dataset (Reddit), while the remainder are smaller datasets. Additionally, FuseGNN even reports that its GCN is slower than DGL for its only mid-size dataset. However, there is no information on whether FuseGNN's GAT is faster than DGL's GAT, as DGL runs out of memory. Seastar also reports that DGL runs out of memory on GAT. 

Though sampling in GNN~\cite{kaler2022accelerating, yang2022gnnlab, waleffe2023mariusgnn, liu2021bgl,lin2020pagraph,song2022rethinking,jangda2021accelerating,ye2023sparsetir} uses large datasets to evaluate training runtime performance, it should be noted that they {sample} the dataset in each iteration to generate a smaller graph to run training. E.g., such works only pick 256 vertices to {sample} the graph, while picking only 10 (for example) neighbors, if those vertices have more than those neighbors. Hence, the resultant {sampled} graph is small enough to be impacted by the framework overhead pitfall. We next show measurements to back up the claim.

\subsection{EVAL-P2: Framework Overhead} \label{sec.overhead}
We now present our proposed methodology, simple yet effective, to measure the framework overhead. We also evaluate to understand the overhead when graph size is varied. We reason for the differences in framework overhead. 
We finally discuss {sampling} in GNN.

\noindent \textbf{A Novel Methodology.}
Eq.~\eqref{eq-overhead} defines the overall framework overhead, which is a sum of the time consumed by the CPU for each training epoch ($k$) in a single-GPU system. 
Kindly note that GPU kernel invocation is asynchronous to the CPU, and it does not wait for the kernel execution to finish-- a practice followed by all GNN systems. Hence, measuring the time taken by each epoch without waiting for the kernels to finish execution roughly measures Python and C++ code runtime executed in the CPU. We can use this time to calculate the overhead. 
In this measurement, $k$ is set to [0, 200). 

\vspace{-6pt}
\begin{equation}\label{eq-overhead}
\textit{Framework$\_$Overhead} = \sum_{k}\textit{$CPU$\_$Runtime$}
\end{equation}
\vspace{-6pt}

\begin{figure}[b]
  \includegraphics[scale= 0.55]{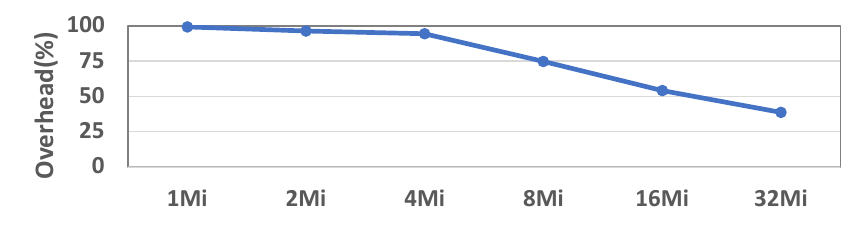}
  \vspace{-12pt}
  \caption{\small DGL overhead ratio while training GCN for 200 epochs on graphs (vertex count = 32,768), where the X-axis is showing edge counts: Almost 100\% overhead when edge-count is $<$ 4 Million.}
  \label{exp-obs-overhead}
\end{figure}




\noindent \textbf{Measurements.}
We first plot the overhead ratio for DGL in Fig.~\ref{exp-obs-overhead} for different edge counts in the graph. Indeed, the overhead is close to 100\% for graphs with an edge count of up to 4 million edges before dropping slowly. For Reddit, DGL has only 5.42\% overhead for GCN.  
This overhead difference \textit{proves that the proposed methodology to measure framework overhead is valid}. We also verified it with other GNN models.



\begin{figure}[t]
  \includegraphics[scale= 0.42]{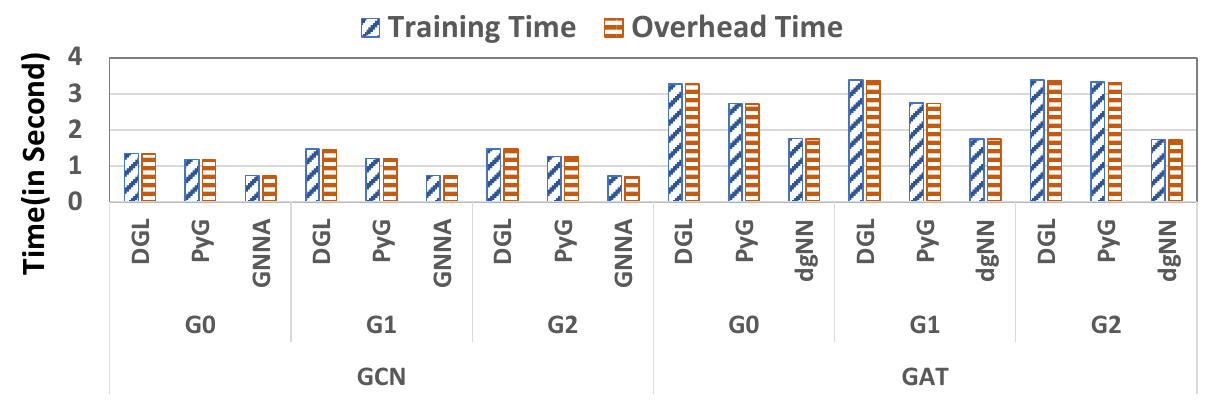}
  \vspace{-24pt}
  \caption{\small Training time and framework overhead for DGL, PyG, GNNA (stands for GNNAdvisor), and dgNN for GNN models training for 200 epochs. We also measured Seastar and have almost 100\% overhead for these datasets.}
  \label{fig-obs-overhead}
\end{figure}

We evaluate the framework overhead for multiple GNN models across three widely used small datasets: Cora, Pubmed, and Citeseer. Fig.~\ref{fig-obs-overhead} shows that the overall training time is dominated by the framework overhead, and not by the kernel performance. The source of overhead varies in different systems. Kindly note that the training loop and model layer (written in Python) remain almost identical between DGL and PyG while the Kernel module is different. Hence they produce slightly different overhead.

In GNNAdvisor,  two changes allow it to have a lower overhead than DGL: removal of bias parameter GCN layers, and use of C++ to write GNN layers instead of Python. These two factors contribute to lowering framework overhead. Though GNNAdvisor also suffers from accuracy issues.
Indeed, many systems develop GNN from scratch and hardly rely on integration to DGL, and can easily outperform it. A new specialized system never suffers the overhead of DGL, which is due to the generality that it offers: a generic message passing infrastructure that can model almost any sparse kernel, conversion of Pytorch tensor to DLPack tensor to DGL-specific tensor are a few things that allow DGL to support newer GNN models, and many general-purpose DL frameworks, such as Pytorch, Tensorflow, MxNet, etc. but also introduces overhead.

On the other hand, systems integrated into DGL, such as GE-SpMM~\cite{gespmmsc2020} (not shown in Fig.~\ref{fig-obs-overhead}), show hardly any performance improvement in their paper for GCN for smaller datasets on which it relied exclusively in its paper despite showing better gSpMM kernel time than Cusparse. This is because the overhead remained the same as DGL.
However, it wrongly attributes this observation to the number of prediction classes not being a multiple of 32. 




\noindent \textbf{Sampling.}
A few GNN systems rely on data sampling to train the GNN model. 
In Reddit and Ogb-product datasets, our measurements for GraphSage sampling using DGL show that the maximum edge count is always less than 0.45 million which is very small. Hence, further measurements show close to 100\% overhead for training on these two datasets. For example, 
the actual training procedure took around 48 seconds in the OGB-Product dataset, while the total overhead during backward and forward computation was also around 48 seconds, showing that overhead during training is close to 100\%. When doing the same measurement in MariusGNN~\cite{waleffe2023mariusgnn}, the training time improved to around 22 seconds but still had close to 100\% overhead. In summary, GPU kernels play no role in training time in this case.

Even for very large datasets, such as paper100M, which is one of the largest datasets used in sampled GNN training, the generated sampled graph has between 13k--2M edges, as reported by MariusGNN.
Fig.~\ref{exp-obs-overhead} shows 100\% overhead for GCN if the edge count is less than 4 million in the sampled graph. 
As GraphSage uses GCN for training after sampling, we conclude that the framework overhead is around 100\% of training time in a sample-based single-GPU GNN system.


\subsection{Memory Consumption Investigation} \label{sec.mem}

DGL and PyG have already addressed the system design pitfalls (\cref{sec.semantic}) even before they appeared in other works, as they were proposed several years ago. However, the solutions are memory-intensive resulting in frequent out-of-memory (OOM) on slightly older GPUs for mid-size datasets. 
It is the first time that excessive memory consumption has forced prior GNN systems to heavily or even exclusively rely on smaller datasets for comparing training runtime which led to EVAL-P2 pitfall (\cref{sec.overhead}). In this section, we investigate the reason behind memory consumption and discover two critical pitfalls which unlike other pitfalls are not related to correctness but are more about \textit{system design efficiency}.



\subsubsection{EVAL-P3: Ignore a Key Baseline for gSpMMve$^T$} 
Our investigation of the excessive memory consumption issue has led to the discovery of a crucial observation in class A GNN: gSpMMve$^T$ API is natively available in Cusparse, yet prior GNN systems, including those that rely on Cusparse, have not used this API to solve SYS-P2. Rather, prior works have introduced additional mechanisms for its implementation without evaluating whether they perform better than the native API of CuSparse.   
Our measurements (Fig.~\ref{exp-gpu-gspmmve}) show that native gSpMMve$^T$ by Cusparse is still the fastest. 
Thus, prior works have introduced an inefficient system design against the key baseline kernel of Cusparse, a widely used library for benchmarking other kernels.

Even worse, as we explain next, such additional mechanisms are the main reason for huge memory consumption and slowdown.
We first analyze DGL for its memory-intensive gSpMMve$^T$ approach followed by a discussion of alternate mechanisms. 
This also serves as an understudy for our reference system discussed in \cref{sec.baseline}.

\noindent $\bullet$ \textbf{DGL.}
To support gSpMMve$^T$, DGL relies on a separate kernel, termed as \textit{eShuffle}, and \textit{edge ID abstraction} in the graph storage.  Every edge of the graph in DGL is assigned a unique numerical identifier in the range of [0, $|E|$). Its allocation is determined by its COO format (used for gSDDMM implementation). 
That is, the offset of each edge in the COO is its implicit edge ID (see Fig.~\ref{fig-format}c). However, the order of edges in DGL's COO (i.e., storage layout) is determined by the dataset file or parsing module which is often performed by the user. Hence, edge ID allocation in DGL is determined by external factors as COO is neither stored in CSR-style (rows are laid out consecutively) nor in CSC-style (columns are laid out consecutively). 
In such a case, both the CSR and the CSC contain an explicit array of edge IDs (Fig.~\ref{fig-format}(d and e)). 

Memory consumption in DGL for storage formats is $2|V|+6|E|$ due to CSR $(|V|+|E|+|E|)$, CSC$(|V|+|E|+|E|)$, and COO($|2E|$) cost. 
Moreover, there is more consumption due to the way kernels are implemented in DGL as explicit edge ID breaks the one-to-one relationship between the non-zero element and the edge-level attention tensor in both its forward and backward computations as discussed next.


DGL relies on \textit{eShuffle}, an internal GPU kernel, to re-arrange the input edge-level tensors using the edge ID array to generate a new edge-level tensor, so that it can use the same forward SpMM API of Cusparse for both gSpMMve and gSpMMve$^T$. That is, the \textit{DGL uses the eShuffle operation during both the forward and backward computation}, and the usage of the edge ID array of CSR or CSC determines whether it is gSpMMve or gSpMMve$^T$ respectively. 

The eShuffle slows down performance by 64\% of Cusparse gSpMM on Reddit (\cref{sec.exp}), and hence is very costly.
Thus, DGL solved the transpose requirement (SYS-P2) by additional memory allocation and slowing down both the forward and backward gSpMM through eShuffle. Due to the eShuffle kernel, DGL also consumes an additional $|E|$ memory per gSpMM.
Hence, any system integrated into DGL suffers from similar issues, but an independently developed system can get additional performance benefits if it suffers from missing transpose (SYS-P2) pitfalls, while forward-focused systems automatically gain 64\% speedup without any optimization. 

\noindent $\bullet$
\textbf{Other Mechanisms.}
Seastar and FeatGraph also rely on edge ID indirection for GAT in both the forward and backward computations. 
FuseGNN relies on \textit{CusparseCsr2cscEx2}() API (which transposes the topology and edge-level tensor) plus their custom forward gSpMMve to implement gSpMMve$^T$. This Cusparse API is slower by 130\% of Cusparse gSpMM on Reddit, thus this mechanism is even more costly.
Others~\cite{huang2021understanding,fu2022tlpgnn,wang2023tc} have mistakenly avoided transpose. 

Out-of-memory in PyG is even more frequent as it cannot run on an A100 GPU with 40GB memory.
This is due to its original design of using the COO format that implements gSpMM by materializing messages at the edge level using global memory before gathering at the destination vertices. 
Lately, PyG has started using CSR without materializing edge-level tensors on GCN, however, for GAT (which uses gSpMMve$^T$), it still materializes them.


\subsubsection{Requirement Mismatch for gSpMMv} \label{sec.mem.gcn}
The class B GNN, (e.g., GCN), does not involve an edge-level tensor, and hence edge ID, COO format, and eShuffle kernels are not needed. Yet, DGL consumes significant memory as it still keeps the same storage, introduces requirement mismatch,  and uses other inefficient designs as discussed next. 

\textit{a)} Cusparse provides only gSpMMve API which is not suited directly as {GCN} needs gSpMMv API. Hence, DGL uses a \textit{dummy edge-level tensor} with each value as 1.0 to use Cusparse API, consuming an additional $|E|$ memory. A few GCN systems~\cite{gespmmsc2020,featgraphsc20,huang2021understanding} are still focused on optimizing gSpMMve despite the focus on GCN. 
\textit{b)} DGL does \textit{normalization by degree} out-of-place. This not only allocates $|V*K|$ sized additional memory(K is feature-length) but the division also becomes a node in the computation graph (DAG) of GNN. Therefore, gradients are also generated and materialized between gSpMM and normalization by degree kernel.
\textit{c)} DGL processes the degree array before each call to normalization by degree due to the presence of empty rows in the sparse matrix. In this case, DGL calls (\textit{clamp}()) API which sets the degree to 1.0 for empty rows for each iteration, introducing further slowdown and additional $|V|$ sized memory.

These design choices are responsible for up to 10$\times$ memory consumption (Fig.~\ref{exp-mem}) and a slowdown in DGL. More importantly, the design decisions have introduced a requirement mismatch: a) using gSpMMve + a dummy edge-level tensor instead of gSpMMv; and b) CusparseSpMM() API supports native gspMMve$^T$, thereby questioning the need for edge ID abstraction and eShuffle design in DGL. 

\subsection{Discussion}



\noindent
\textbf{Fundamental Problem or Evaluation Oversight?}
This is an important question because an academic prototype does not go through the rigorous evaluation as a commercially developed code. However, in DL training where runtime and memory consumption are the two most important metrics, the two pitfalls    
have questioned our understanding of GNN system design and evaluation. Indeed, we show that many of the current systems have performance regression if mid-size datasets (for EVAL-P2) or the right baseline (for EVAL-P3) are chosen (\cref{sec.exp.kernel}). Hence, we cannot term these pitfalls as evaluation oversight. 

\noindent
\textbf{Impact: Applicability of Optimization Techniques.}
Pitfalls do impact performance, contradicting the published results thereby questioning the applicability of some of the optimization techniques proposed earlier (\cref{sec.exp.kernel}). 
However, that is not the only impact we want to emphasize. Our main objective is to highlight a fundamental flaw in the evaluation and design process that the community may not be aware of thereby we should not remain trapped in such errors in the future thereby compromising the optimization process. 
For example, we show that the proposed reference system ({\graphpy}) has slower end-to-end training time than a pitfall-manifested system on smaller datasets despite having better runtime for underlying kernels. Such results are unlikely to impress authors themselves and reviewers in a conference or journal unless pitfalls are known.
It also has huge repercussions for future systems research as outlined in \cref{sec.analyze} (first bullet point).

\noindent \textbf{Impact: Broken GNN Workflow.} 
The study suggests that we have masked the true comparison of underlying optimization techniques while measuring training time as evaluation on smaller datasets can not be relied upon due to framework overhead. Indeed, we along with others~\cite{fu2022tlpgnn} have noticed that GNNAdvisor cannot be executed for mid-size datasets at all for GCN, showing the broken GNN workflow.
Further, scant or inconclusive results on mid-size datasets coupled with accuracy issues together question the whole GNN workflow: the benchmarking practices, the drawn conclusions, and lessons learned in many prior works.

\noindent \textbf{Hypothesis:}
For EVAL-P2, we believe the OOM condition by DGL and PyG did not allow conclusive comparison on mid-size labeled datasets, thereby many prior works rely on smaller datasets. 
Further, earlier versions of GPUs had less memory on which prior works had to rely leading to OOM by baselines. 
Moreover, access to limited choices of labeled datasets and no significant precedent of using synthetic datasets forced us to rely excessively on smaller datasets, thereby leading to a possible compromise on system evaluation.

\noindent \textbf{Recommendation \#3.}
It is aptly clear that smaller datasets alone cannot be used to conclude better runtime performance as it simply measures the framework overhead. 
To negate the impact of overhead, end-to-end performance measurements should also focus on mid-size datasets for evaluation.

\noindent \textbf{Recommendation \#4.}
Once system accuracy measurement confirms the correctness of the system design, but limited labeled datasets do not allow us to verify the runtime performance, we recommend relying on datasets through generation tools~\cite{graph500,gt_graph} and other graph repositories~\cite{matrix_market,snap} for training runtime comparison only.

\noindent \textbf{Recommendation \#5.}
Frequent OOM should be analyzed (as we do in \cref{sec.mem}) and presented with the same zeal as the runtime rather than blindly assuming that state tensors are responsible for memory consumption. Otherwise, we continue to have limited choices for current GNN systems to evaluate training performance on mid-size datasets.

\section{Future Direction and A Reference System} \label{sec.analyze}

After identifying pitfalls, the natural progression leads us to the question of overcoming these pitfalls systematically beyond recommendations. We present three important directions, as discussed next.

\noindent $\bullet$
\textbf{Framework Overhead Research.} The advent of framework overhead, though can be measured, presents the community with a new area of research. Specifically, when we zoom into one kernel operation in a DL model, it is becoming faster due to powerful GPUs and better algorithms along with other innovations such as quantization, pruning, \textit{sampling}, etc. At the same time, framework overhead remains constant and understudied, which runs in a single CPU core. Hence, we infer that the DL framework is now becoming more like an Operating System whose job is to allow access to hardware without introducing much overhead. Thus, the future lies in studying those overheads.

\noindent $\bullet$ 
\textbf{Invasive Effort.} Some pitfalls can be addressed via only invasive efforts by prior works themselves-- e.g., through developing backward computation code, fixing the order in backward computations, including bias and other mistakenly removed operations to model layers, etc. while using accuracy measure as feedback.

\noindent $\bullet$ 
\textbf{Reference System.}
Current single-GPU systems suffer from many pitfalls, as we discussed so far, that forbid their usage as a good end-to-end baseline. We argue the need for a new reference system to serve as a good baseline for promoting future systems research. 
Further, the reference system is used to show that pitfalls can stifle future innovations.

\subsection{{\graphpy}: A Reference System} \label{sec.baseline}

We present {\graphpy}, a novel single-GPU GNN system rooted in clear requirement understanding.  
The proposed system is also targeted to ensure \textit{practicality} so that some of those ideas can be integrated into prior works productively to solve a few system design pitfalls. 
\textit{Efficiency} is another goal to save memory and not slow down performance, which means that we cannot use the DGL or PyG solution. Kindly note that forward gSpMM optimization is not a goal here.

\begin{figure}[b]
  \centering
  \includegraphics[scale= 0.35,left]{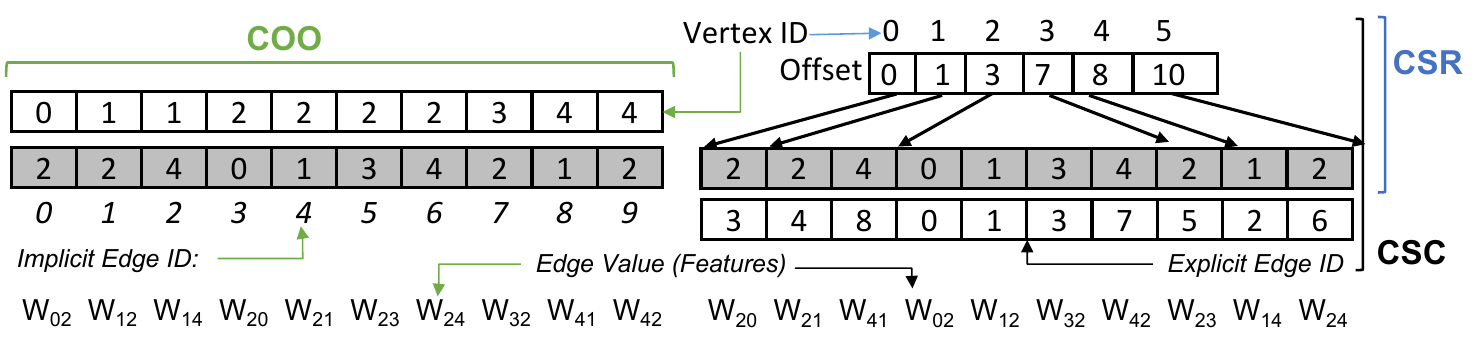}
  \vspace{-18pt}
  \caption{\small The data format layout in {\graphpy} for sample graph of Fig.~\ref{fig-format}(b): There is only one copy of the grayed array representing Column ID array, which is shared among COO, CSR, and CSC. The Offset array is shared between CSR and CSC, while the edge ID array is specific to CSC only.}
  \label{fig-graphpy-format}
\end{figure}

\subsubsection{Storage Format: Symmetry \& Edge ID Order}

Kindly note that GNN datasets are symmetric, where each edge is stored in both directions. We \textit{exploit the symmetry} where we keep one copy of the topology that serves both CSR and CSC. Further, our COO format is arranged in CSR-style (i.e., rows are laid out sequentially) as shown in Fig.~\ref{fig-graphpy-format}. This means that the COO and CSR share their column ID array (grayed part in Fig.~\ref{fig-graphpy-format}) 
However, the edge-level features are generated individually for each edge using trainable parameters and are not symmetric. 
Hence, edge IDs may still be required to be stored for both storage formats.
To this end, we propose a \textit{novel edge ID reordering}, where CSR and COO (which is stored in CSR-style) both are allocated consecutive edge IDs. Hence, we do not keep an explicit edge ID array for them.
However, CSC requires an explicit edge ID array as symmetry in topology does not imply symmetry in edge ID. This does introduce heterogeneity in the graph storage. 


In total, the format costs in {\graphpy} is only $|V| + 3|E|$ where CSR/CSC needs $(|V| + |E|)$ + COO ($|E|$) + Edge ID array ($|E|$) in CSC  for class A GNN. For GCN, {\graphpy} need only $(|V| + |E|)$. 


 \subsubsection{Kernel Design}

\noindent \textbf{{gSpMMve$^T$}.}
{\graphpy} implements $gSpMMve^T$ natively as one single operation without a separate eShuffle kernel. It uses edge ID to fetch the correct edge-tensor values while keeping the rest of the gSpMMve code the same. 
This avoids the additional memory that would have been required, while also improving the performance compared to DGL. 
Other works can easily implement such an approach, even if they rely on a custom storage format as the sequential edge ID will still be applicable for forward gSpMMve without requiring any change, while format-specific but explicit edge ID array needs to be generated for $gSpMMve^T$. 
Kindly note that optimizing gSpMMve is not within the scope of this work.

\noindent \textbf{gSpMMv$^T$}
For backward computation, GCN requires transpose of the topology only, as there is no edge-level tensor present in GCN. As GNN datasets are symmetric, it does not require any additional operation as  gSpMMv$^T$ is equivalent to gSpMMv.
{\graphpy} supports \textit{gSpMMv} API natively where the edge-level tensor is not an input.
The API also accepts an optional boolean flag to indicate normalization by degree flag. When the flag is true, it only normalizes the result of gSpMM as part of the same kernel. However, in the backward computation, a separate \textit{in-place} normalization kernel is called before calling backward gSpMMv.




This kernel definition, though very intuitive 
provides substantial memory saving to {\graphpy} on two accounts. (1) Compared to Cusparse compliant gSpMMve, {\graphpy} never need to allocate dummy edge-level tensor saving O($|E|$) memory. This also results in a substantial performance gain. 
(2) Even in the case of a separate normalization, {\graphpy} uses the same tensor as input and output both by using in-place normalization inside {\graphpy}. 

\noindent \textbf{gSDDMM and New Avenue of Data Locality.} 
This kernel operates on the vertex-level features of rows and columns for each edge, and generates output at edge-level. 
The COO format stored in CSR-style leads to a new avenue of data locality in gSDDMM, which is purely enabled due to the proposed edge ID reordering while providing a fully workload-balanced gSDDMM solution. Specifically, we first divide the COO format equally among warps of the GPU to achieve workload balance. Then within a warp, many edges have the same row ID as COO is laid out in CSR-style.
Hence, the vertex-level features of the same row ID across edges can be fetched once and \textit{reused} multiple times within the warp. 

In contrast, prior gSDDMM solutions can either achieve workload balance or data reuse, but not both.
For example, DGL, which uses COO, can achieve workload balance, but can not reuse the features of row ID as its COO is not laid out in CSR-style.  
On the other hand, a CSR-based gSDDMM offers data locality for vertex-features of rows, but cannot offer workload balancing. Hence, \textit{all prior systems have introduced a  trade-off: the workload-balanced COO-gSDDMM or data-locality based CSR-gSDDMM.}

\begin{table}[t]
\footnotesize
\caption{\small Dataset details. * indicates labeled dataset, while the rest use 150 generated features and 7 prediction classes.} 
\centering 
\begin{tabular}{l l r r r r} 
    \hline\hline 
    Graph & Vertex & Edge & Feature & Predict\\ [0.5ex] 
    Dataset & Count & Count & Length & Class\\ 
    \hline 
    Cora(G0)* & 2,708 &  10,858 & 1,433 & 7\\ 
    Citeseer(G1)*  & 3,327 & 9,104 & 3,703 & 6\\
    Pubmed(G2)* & 19,717 &  88,648 & 500 & 3\\
    Amazon(G3) & 400,727 & 6,400,880 & 150 & 7\\ 
    As-Skitter(G4) & 1,696,415 & 22,190,596 & 150 & 7\\
    Cit-Patent(G5) & 3,774,768 & 33,037,894 & 150 & 7 \\ 
    Stackoverflow(G6) & 2,601,977 &	95,806,532 & 150 & 7\\ 
    Hollywood(G7) & 1,069,127 & 112,613,308 & 150 & 7\\
    LiveJournal(G8) & 4,847,571 & 137,987,546 & 150 & 7 \\
    OGB-Prduct(G9)* & 2,449,029	& 123,718,280 & 100 & 47 \\
    Reddit(G10)* & 232,965 & 229,231,784 & 602 & 41 \\
    Orkut(G11)* & 3,072,627 & 234,370,166 & 150 & 7\\
    UK-2002(G12) & 18,520,486 & 596,227,524 & 150 & 7 \\
    Kron-25(G13) & 33,554,432 & 1,073,741,824 & 150 & 7 \\
    \hline 
\end{tabular}
\label{table-dataset} 
\end{table}

\section{Experiments} \label{sec.exp}
We already evaluated many prior works throughout the paper. We now evaluate {\graphpy} with DGL on end-to-end training time and memory consumption. We also compare against GNNAdvisor~\cite{wang2021gnnadvisor}, Huang et al~\cite{huang2021understanding},  GE-SpMM~\cite{gespmmsc2020}, and FeatGraph~\cite{featgraphsc20} using the recommendation, where we found that CUDA API level comparison may be apt due to the absence of backward computation or wrong accuracy.

The datasets for the experiments are listed in Table~\ref{table-dataset}. Only Cora, Pubmed, Citeseer, Reddit, and OGB-product datasets are labeled. The first three datasets are referred to as small. The experiments are run on  Nvidia A100 GPU, which has 40GB of memory.
We focus on node classification using GCN, GIN, and GAT (with 1 and 3 heads implying edge feature length). In GCN and GAT, the intermediate feature length is 16, which implies that GAT-3 has 48 as the feature length for each vertex. For GIN, the intermediate vertex feature length is 64. The feature lengths have been set based on the code of the original model paper, thus covering various feature lengths.

\begin{figure}[b]
  \includegraphics[scale= 0.51]{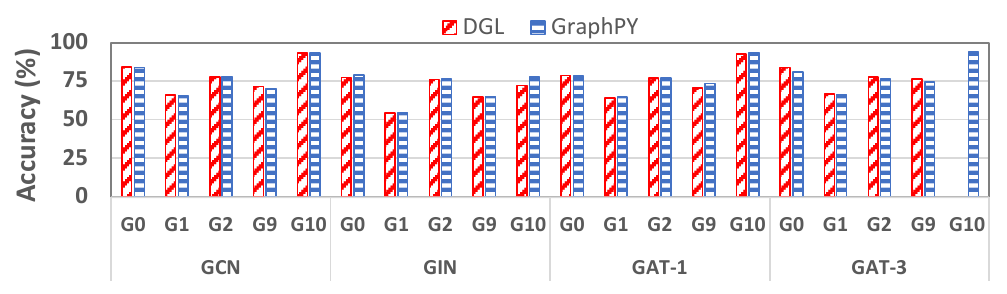}
  \vspace{-18pt}
  \caption{\small Accuracy comparison of {\graphpy} and DGL. For other systems, please refer back to Fig.~\ref{pitfall-accuracy} and \cref{sec.accuracy}.} 
  \label{exp-accuracy}
\end{figure}


\noindent \textbf{GNN Traning Accuracy}
Fig.~\ref{exp-accuracy} shows the accuracy of {\graphpy} compared to DGL to clearly outline that {\graphpy} matches the accuracy almost exactly.

\subsection{GNN Training}
This section compares {\graphpy} with DGL as PyG runs out of memory on Reddit and OGB-Product. 

\begin{figure}[t]
  \includegraphics[scale= 0.48]{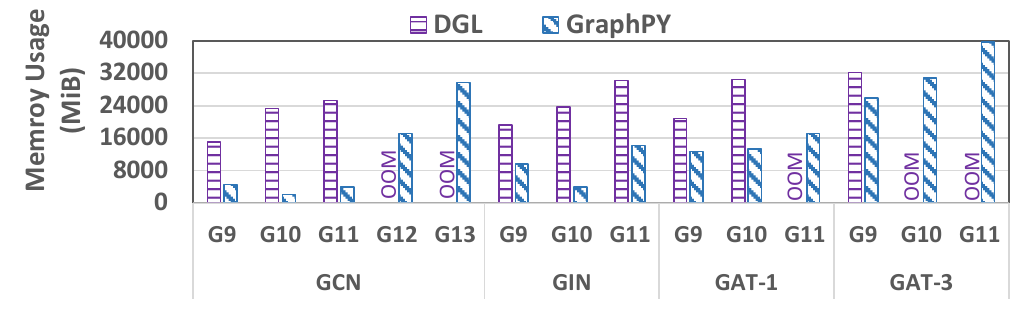}
  \vspace{-6pt}
  \caption{\small Memory consumption comparison on various models. OOM refers to an out-of-memory condition. (Lower is better).}
  \label{exp-mem}
\end{figure}

\noindent \textbf{Memory Consumption}
Fig.~\ref{exp-mem} shows that {\graphpy} reduces memory consumption on average by 6.92$\times$ on GCN, 3.4$\times$ on GIN, and 1.96$\times$ on GAT-1 models in comparison to DGL.
E.g., on Reddit (G10), {\graphpy} (DGL) consumes only 2.1GB (23.2GB), 4.0GB (23.7GB), 13.3GB (30.3GB), and 30.9GB(OOM) on GCN, GIN, GAT-1, and GAT-3 respectively.
We exclude results for G3--G8 in Fig.~\ref{exp-mem} as the focus is to study the out-of-memory behavior, though we observe a similar pattern for those datasets.


The additional storage requirement, and memory allocation for the dummy edge-level tensor in GCN and GIN and for intermediate eShuffle results in GAT leads to huge memory consumption in DGL.
(1) Results reveal that GCN and GIN memory consumption is also proportional to $|E|$ in DGL due to the mismatch, while {\graphpy} closely follows $|V|$.
(2) For GAT-3, DGL has its own implementation as CuSparse has no support for the vector edge feature. However, it consumes more memory and runs out of memory on Reddit. (3) Because of the DGL-induced memory overhead, any system integrated into DGL, such as Ge-SpMM and FeatGraph cannot lower its memory requirement, as their focus is only on basic kernel.







In summary, devising solutions rooted in the \textit{requirement understanding} has the potential to lower memory consumption.
Another important takeaway is that \textit{{\graphpy} can train GCN on a billion-edge graph (G13: Kron-25), a first on a GPU}, consuming 29.8 GB only, while DGL cannot even run a graph with around half a billion-edge (G12: UK-2002) when using the same feature-count (150) and prediction classes (7).

\begin{figure}[b]
  \includegraphics[scale= 0.49,right]{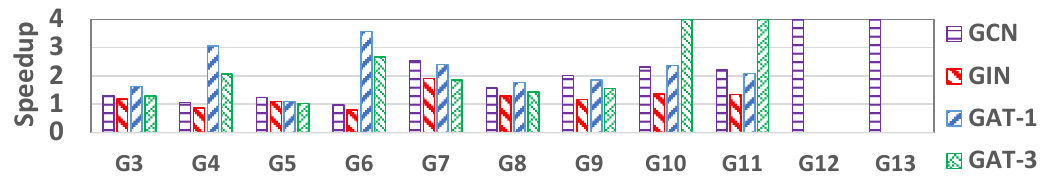}
  \vspace{-18pt}
  \caption{\small {\graphpy} speedup over DGL for GNN training time (200 epochs) in GPU for mid-size datasets. A speedup of 4 means that DGL has OOM while {\graphpy} runs. (higher is better)}
  \label{exp-gnn-gpu-large}
\end{figure}

\noindent \textbf{Training Time.}
Fig.~\ref{exp-gnn-gpu-large} shows that {\graphpy} performs much better than DGL for all datasets while training various GNN models. 
For Reddit, {\graphpy} achieves around 2.30$\times$ speedup for GCN, 1.36$\times$ for GIN, and 2.37$\times$ for GAT-1 over DGL, while for GAT-3 DGL resulted in OOM. We do not show results on smaller datasets, as it is a comparison of framework overhead due to which {\graphpy} performs better.

\begin{figure}[t]
  \centering
  \includegraphics[scale= 0.52]{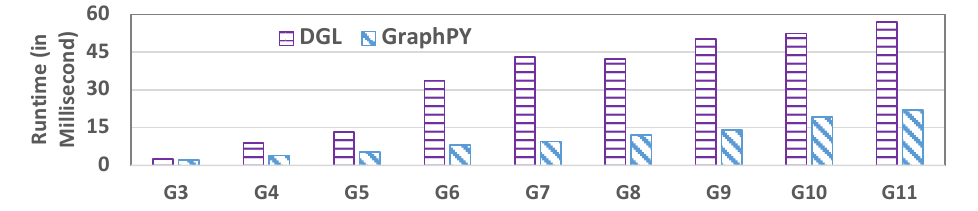}
  \vspace{-12pt}
  \caption{\small gSDDMM kernel runtime for DGL and {\graphpy} for feature-length 32 (lower is better).}
  \label{exp-sddmm}
\end{figure}

The performance improvement is purely based on designing kernels as per GNN requirements. 
Specifically, usage of CSR-style COO increases the data locality in gSDDMM, resulting in a huge speedup as shown in Fig.~\ref{exp-sddmm}, which was inspired by finding the solution for the transpose of the sparse matrix.
This resulted in an average 2.99$\times$ speedup. Many other GNN systems do not have a separate gSDDMM. FeatGraph does provide gSDDMM, but its idea has been merged with DGL. Cusparse is very slow~\cite{ye2023sparsetir}, which we have also observed, and hence has not been plotted.

We discuss gSpMM variants in more detail in \cref{sec.exp.kernel} to show how those kernels are also responsible for better performance.

\subsection{Pitfalls and Performance Evaluation} \label{sec.exp.kernel}

We have already shown how pitfalls have impacted accuracy in \cref{sec.accuracy} and Fig.~\ref{pitfall-accuracy}. This section shows that pitfalls have masked proper run-time evaluations and can stifle future innovations if they only focus on smaller datasets or have other system design and evaluation pitfalls. We use a mix of kernel time and training time (\cref{exp.kernel.pitfall}) to show how pitfalls impact runtime.




\subsubsection{Quantitative Pitfall Discussion via gSpMMve$^T$}

The gSpMMve$^T$ is natively implemented by Cusparse, which serves as the baseline. However, DGL does not rely on this native API but rather uses eShuffle + gSpMMve of Cusparse. 
The cost of eShuffle is derived by subtracting Cusparse gSpMMve time from the DGL runtime of this kernel and is verified using our eShuffle implementation. 
For other systems that do not provide native gSpMMve$^T$, we measure their gSMMve time and add our eShuffle to estimate their gSpMMve$^T$ time. Other mechanisms such as the usage of CuSparseCsr2cscEx2() API are not considered as it is almost 2$\times$ more costly than eShuffle.
GNNAdvisor is not plotted as it does not provide gSpMMve.

\begin{figure}[t]
  \includegraphics[scale= 0.60]{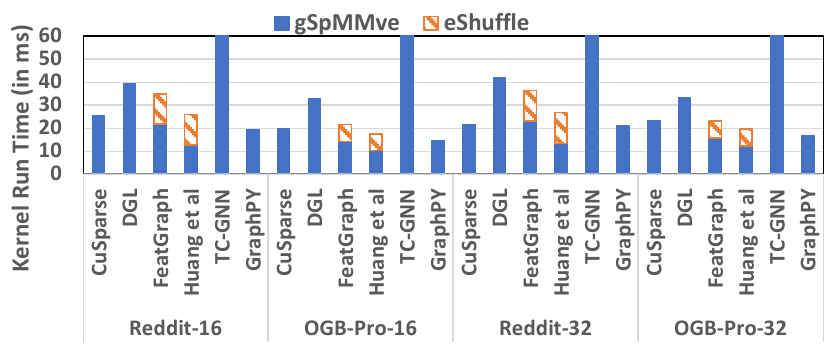}
  \vspace{-24pt}
  \caption{\small gSpMMve$^T$ kernel time comparison for 16 and 32 feature length (Lower is better): Cusparse provides native support for gSpMMve$^T$. TC-GNN is very slow and has been clipped. {\graphpy} is discussed in \cref{exp.kernel.graphpy}.}
  \label{exp-gpu-gspmmve}
\end{figure}

The results (Fig.~\ref{exp-gpu-gspmmve}) clearly show that  DGL, TC-GNN, and FeatGraph are slower than the Cusparse result by a big margin, while Huang et al. is very close. 
a) TC-GNN  kernel is slower by over 3$\times$ and 54$\times$ on two popular mid-level datasets of OGB-Product and Reddit respectively for feature-length 32. \textit{These results prove that focusing only on smaller datasets (EVAL-P2) is a poor benchmarking practice.}

b) By ignoring the native transposed gSpMM API of Cusparse, DGL slows down by 1.64$\times$ over to Cusparse. Worst, DGL also slows down its forward computation as it uses eShuffle there. \textit{This proves the importance of requirement understanding and evaluation (EVAL-P3) in system building.}

c) The performance gain by Huang et al. is not noticeable for this kernel and is also slower in some cases. Rather, eShuffle time is nearly equal to gSpMMve time by Huang et al. in Reddit. \textit{This proves that emulated forward-only evaluation (SYS-P1) is not a substitute for GNN training measurement. These insights have been unearthed due to this paper.}

\subsubsection{Quantitative Pitfall Discussion via gSpMMv} 
%
A similar analysis for gSpMMv is plotted in Fig.~\ref{exp-gpu-gspmmv}.  We use Cusparse as the baseline.
Kindly note that GCN is a simple model, where this kernel dominates the run-time unless pitfalls hide it. Hence, all single-GPU GCN systems have mostly focused on kernel-level optimizations.  
a) TC-GNN (published in 2023) is the slowest system. It is slower than Cusparse by 3.07$\times$ and 38.67$\times$ on the mid-size datasets of Ogb-Product and Reddit respectively. 
b) GNNAdvisor is the next slowest system despite using a workload-balanced technique (neighbor grouping). It is slower by 1.20$\times$ than Cusparse. It is slower by 1.98$\times$ than the workload-imbalanced solution of Ge-SpMM, even though it uses its caching design.
Kindly note that both TC-GNN and GNNAdvisor rely on smaller datasets (EVAL-P2), do not measure accuracy (EVAL-P1), and suffer from many system design pitfalls. \textit{Hence, these results have shown that pitfalls have impacted true performance benchmarking and question the contribution of the optimization technique themselves.}

c) Cusparse and Huang et al provided gSpMMve and not gSpMMv.  Huang et al (despite a workload-balanced solution) still perform similarly to gSpMMv of workload-imbalanced solution of Ge-SpMM. The former offers gspMMve and not gSpMMv thereby incurring a slowdown due to the additional data load of a dummy edge-tensor. \textit{Hence requirement mismatch (\cref{sec.mem.gcn}) also impacts the progress of system research}.




\subsubsection{{\graphpy} Kernel Analysis} \label{exp.kernel.graphpy}

\begin{figure}[t]
  \includegraphics[scale= 0.60]{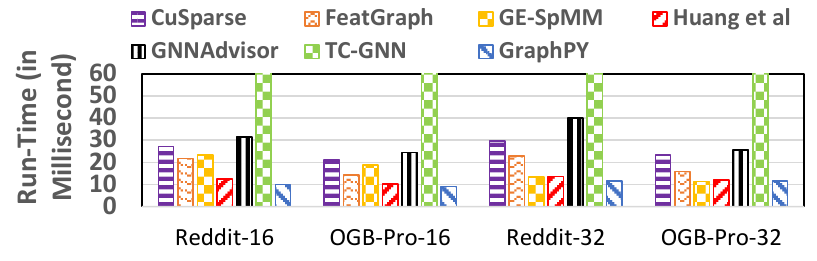}
 \vspace{-6pt}
  \caption{\small gSpMMv GPU kernel runtime (lower is better): TC-GNN runtime has been clipped. {\graphpy} is discussed in \cref{exp.kernel.graphpy}. }
  \label{exp-gpu-gspmmv}
\end{figure}

{\graphpy} kernels are vertex-parallel implementations without any workload balancing technique as the focus is on solving the pitfalls. 
a) For gspMMve$^T${\graphpy} still achieves (Fig.~\ref{exp-gpu-gspmmve}) 32.23$\times$, 1.50$\times$, and 1.16$\times$ speedup compared to TC-GNN, FeatGraph, and Cusparse respectively.
{\graphpy} outperforms the workload-balanced kernel by Huang et al. by 1.16$\times$. The evaluation justifies the proposed additional mechanism for implementing gSpMMve$^T$. 
b) For gSpMMv (Fig.~\ref{exp-gpu-gspmmv}), {\graphpy} is better than all prior works, 
e.g., it outperforms GNNAdvisor's workload-balanced gSpMMv by 2.87$\times$.
Kindly note that we have not included the graph pre-processing which is needed to achieve workload balancing.
Compared to the workload-balanced solution of Huang et al, {\graphpy} performed slightly better. We believe it could have made a greater impact had it implemented a native gSpMMv.

\begin{figure}[h]
  \includegraphics[scale= 0.45]{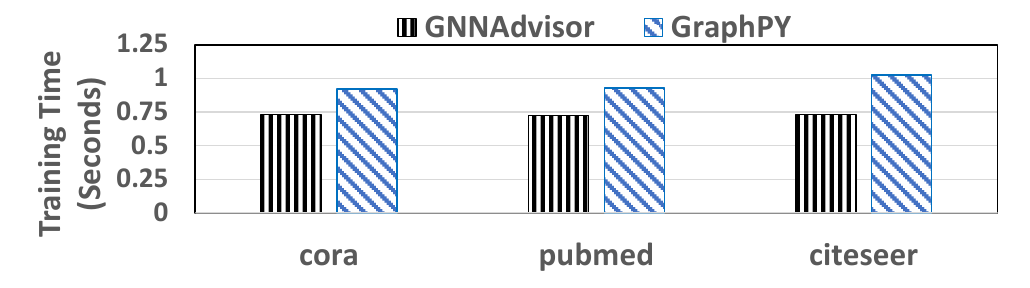}
  \vspace{-12pt}
  \caption{\small {\graphpy} is slower than GNNAdvisor on smaller datasets. This proves how pitfalls can stifle system research. (Lower is better)}
  \label{exp-gpu-graphpy}
\vspace{-12pt}
\end{figure}

\subsubsection{Impact of Pitfalls On Training Time} 
\label{exp.kernel.pitfall}

We pick GNNAdvisor to show that pitfalls can derail future systems research-- {\graphpy} in this example.
We observe that GNNAdvisor cannot train on mid-size datasets due to memory corruption errors as pointed out earlier~\cite{fu2022tlpgnn}.  This forces measurements on smaller datasets of Cora, Pubmed, and Citeseer where {\graphpy} is slower as shown in Fig.~\ref{exp-gpu-graphpy}. E.g., in Pubmed, {\graphpy} takes 0.929 seconds for training while GNNAdvisor takes only 0.722 seconds. 
Note that GNNAdvisor does not have a bias parameter, and wrongly fuses gSpMMv and normalization by degree kernels, 
all of which lowers the framework overhead. In absence of the knowledge about framework overhead pitfalls (in general), and system design pitfalls in GNNAdvisor, advances proposed by {\graphpy} may not stand the scrutiny of authors themselves or reviewers of conferences and journals; thereby stifling future innovations. 

\section{Conclusion} \label{sec.conclusion}
Our in-depth analysis highlighted many critical pitfalls in current single-GPU GNN systems.
We presented their impact, hypotheses, crucial recommendations, and future directions. We also designed a practical and compelling solution to solve some of these pitfalls, 
whose important takeaway is that a careful design rooted in a clear understanding of requirements can reduce memory consumption to the extent that we can train a billion-edge graph in a single GPU.
We hope that our recommendations and {\graphpy} can be used as a state-of-art reference by future GNN research which can be used to genuinely advance the GNN system optimization.

\newpage
 \balance
 {
 \bibliographystyle{refer/abbrv}
 \bibliography{refer/GraphPy}
 }
 
\end{document}